\documentclass[runningheads]{llncs}

\usepackage{eccv}

\usepackage{eccvabbrv}

\usepackage{booktabs}

\usepackage[accsupp]{axessibility}  
\usepackage{multirow}
\usepackage{url}
\usepackage{algorithm}
\usepackage{algorithmic}
\usepackage{amsmath}
\usepackage{upgreek}
\usepackage{amsmath}
\usepackage{graphicx}
\usepackage{upgreek}
\usepackage{amsmath}
\usepackage[utf8]{inputenc} 
\usepackage[T1]{fontenc}    
\usepackage{url}            
\usepackage{booktabs}       
\usepackage{amsfonts}       
\usepackage{nicefrac}       
\usepackage{wrapfig}
\usepackage{microtype}      
\usepackage{xcolor}         
\usepackage{graphicx}
\usepackage{pifont}       
\usepackage{bbding}       
\usepackage{fontawesome}
\newcommand{\hupar}[1]{\noindent\textbf{#1}\quad}

\usepackage[pagebackref,breaklinks,colorlinks,citecolor=eccvblue]{hyperref}

\usepackage{orcidlink}

\begin{document}

\title{Learning to Robustly Reconstruct Dynamic Scenes from Low-light Spike Streams} 

\titlerunning{Reconstruct from Low-light Spike Streams}

\author{Liwen Hu\inst{1} \and
Ziluo Ding\inst{2} \and
Mianzhi Liu\inst{3} \and
Lei Ma\inst{1, 3}\thanks{Corresponding author.} \and
Tiejun	Huang\inst{1}
}

\authorrunning{L. Hu et al.}

\institute{State Key Laboratory of Multimedia Information Processing, School of Computer Science, Peking University 
\email{\{huliwen, lei-ma, tjhuang\}@pku.edu.cn}\\
 \and
Beijing Academy of Artificial Intelligence\\
\email{\{ziluoding\}@baai.ac.cn}
 \and
College of Future Technology, Peking University\\
\email{\{liumianzhi\}@stu.pku.edu.cn}
}

\maketitle

\begin{abstract}

Spike camera with high temporal resolution can fire continuous binary spike streams to record per-pixel light intensity. By using reconstruction methods, the scene details in high-speed scenes can be restored from spike streams. However, existing methods struggle to perform well in low-light environments due to insufficient information in spike streams. To this end, we propose a bidirectional recurrent-based reconstruction framework to better handle such extreme conditions. In more detail, a \textbf{l}ight-\textbf{r}obust \textbf{rep}resentation (LR-Rep) is designed to aggregate temporal information in spike streams. Moreover, a fusion module is used to extract temporal features. Besides, we synthesize a reconstruction dataset for high-speed low-light scenes where light sources are carefully designed to be consistent with reality. The experiment shows the superiority of our method. Importantly, our method also generalizes well to real spike streams. Our project is: \url{https://github.com/Acnext/Learning-to-Robustly-Reconstruct-Dynamic-Scenes-from-Low-light-Spike-Streams/}. 
\end{abstract}
\section{Introduction}

\begin{figure*}[thbp]
\includegraphics[width=\linewidth]{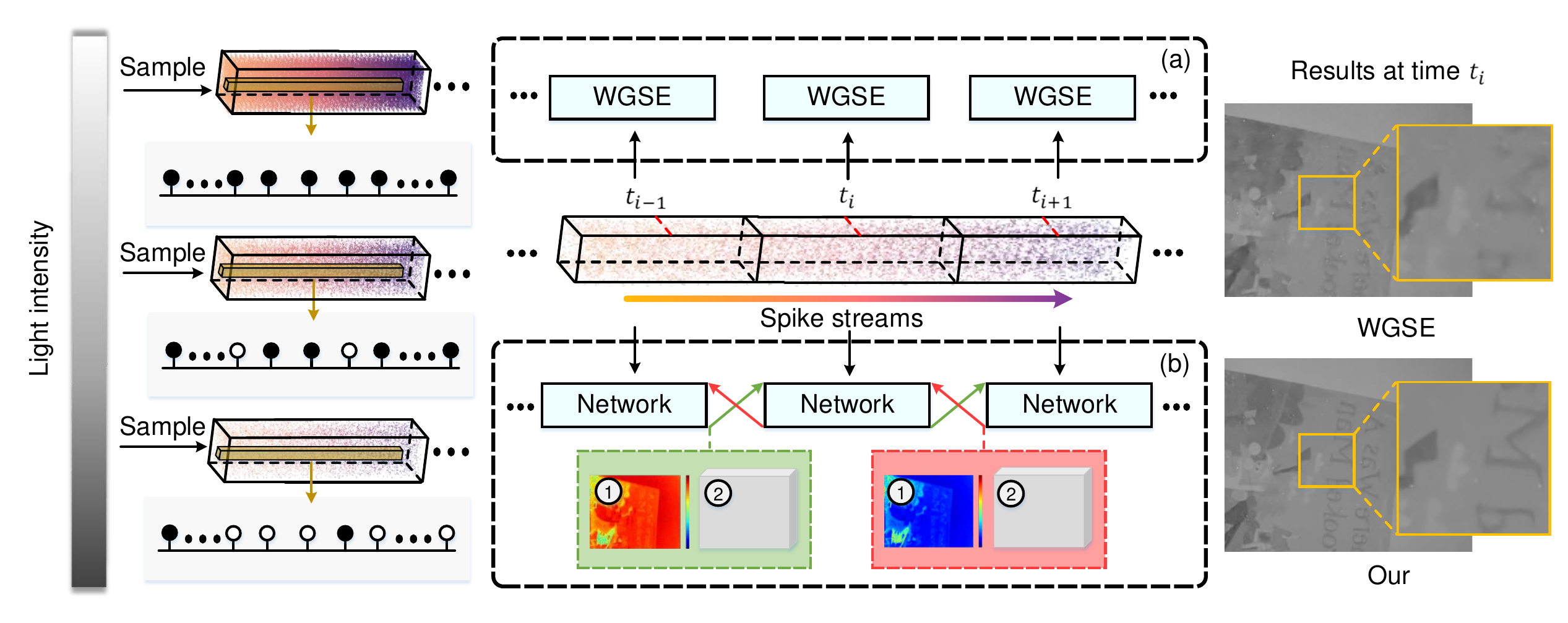}
\centering
\caption{{Overview of reconstruction for high-speed spike streams. \textbf{Left}: with decreasing light intensity, more sparse spike streams are difficult to extract features. A black circle is a spike. 
\textbf{Middle}: (a) The state-of-the-art method, WGSE \cite{rec6}. The arrow with a gradient color is the timeline. (b) Our reconstruction method. Green (red) lines denote the forward (backward) data flow. \ding{172} (\ding{173}) is the release time of spikes (temporal features). \ding{172} (\ding{173}) in forward and backward data flow is independent. \textbf{Right}: reconstructed results from WGSE and our method.}}\label{teasor}
\end{figure*}

As a neuromorphic sensor with high temporal resolution (40,000 Hz), spike camera \cite{spikecamera,huang20221000} has shown enormous potential for high-speed visual tasks, such as reconstruction \cite{rec0,rec1,rec2,rec3,rec4, rec5, dong2024joint, dong2024super, dong2024learning}, optical flow estimation \cite{2022scflow,zhao2022learning,flow2}, and depth estimation \cite{2022spkTransformer,deep1,deep2}. Different from event cameras \cite{dvs1,dvs2,dvs3}, it can record per-pixel light intensity by accumulating photons and firing continuous binary spike streams. Correspondingly, high-speed dynamic scenes can be reconstructed from spike streams. Recently, many deep learning methods have advanced this field and shown great success in reconstructing more detailed scenes. However, existing methods struggle to perform well in low-light high-speed scenes due to insufficient information in spike streams.

A dilemma arises for visual sensors, that is, the quality of sampled data can greatly decrease in a low-light environment \cite{dark_camera0,dark_camera1,dark_camera2,zhao2022learning, graca2023optimal}.
Low-quality data creates many difficulties for all kinds of vision tasks.
Similarly, the reconstruction for the spike camera also suffers from this problem. To improve the performance of reconstruction in low-light high-speed scenes, two non-trivial matters should be carefully considered. First, 
constructing a low-light high-speed {{scene}} dataset for spike camera is crucial to evaluating different methods. However, due to the frame rate limitations of traditional cameras, it is difficult to capture images clearly in real high-speed scenes as supervised signals. Instead of it, a reasonable way is to synthesize datasets for spike camera \cite{rec3, rec4,2022scflow,2022spkTransformer}. To ensure the reliability of the reconstruction dataset, synthetic low-light high-speed scenes should be as consistent as possible with the real world, \eg light source. Second, as shown in Fig.~\ref{teasor}, with the decrease of illuminance in the environment, the total number of spikes in spike streams {{decreases}} greatly which means the valid information in spike streams can greatly decrease. Fig.~\ref{teasor}(a) shows that the state-of-the-art method often fail under low-light conditions since they have no choice but to rely on inadequate information.


In this work, we aim to address all two issues above-mentioned. In more detail, a reconstruction dataset for high-speed low-light scenes is proposed. We carefully design the scene by controlling the type and power of the light source and {{generating}} noisy spike streams based on \cite{zhao2022spikingsim}. 
Besides, we propose a light-robust reconstruction method as shown in Fig.~\ref{teasor}(b). Specifically, to compensate for information deficiencies in low-light spike streams, we propose a \textbf{l}ight-\textbf{r}obust \textbf{rep}resentation (LR-Rep).
In LR-Rep, the release time of forward and backward spikes is used to update a \textbf{g}lobal \textbf{i}nter-\textbf{s}pike \textbf{i}nterval (GISI). Then, to further excavate temporal information in spike streams, LR-Rep is fused with forward (backward) temporal features. During the feature fusion process, we add alignment information to avoid the misalignment of motion from different timestamps. Finally, the scene is clearly reconstructed from fused features.

Empirically, we show the superiority of our reconstruction method. Importantly, our method also generalizes well to real spike streams. In addition, extensive ablation studies demonstrate the effectiveness of each component. The main contributions of this paper can be summarized as follows:


$\bullet$ A reconstruction dataset for high-speed low-light scenes is proposed. We carefully construct varied low-light scenes that are close to reality.


$\bullet$ We propose a bidirectional recurrent-based reconstruction framework where a light-robust representation, LR-Rep, and fusion module can effectively compensate for information deficiencies in low-light spike streams.

$\bullet$ Experimental results on real and synthetic datasets have shown our method can more effectively handle spike streams in high-speed low-light scenes than previous methods.

\section{Related Work}
\subsection{Low-light Vision}
Low-light environment has always been a challenge not only for human perception but also for computer vision methods. For \textbf{traditional cameras}, some works \cite{wei2018RetinexNet, jiang2021enlightengan, dark_camera0, wu2023LLIE, xu2023LLIE, Cai2023LLIE, wang2023LLIE, Fu_2023LLIE} mainly concern the enhancement of low-light images. Wei \textit{et al.} \cite{wei2018RetinexNet} propose the LOL dataset containing low/normal-light image pairs and propose a deep Retinex-Net including a Decom-Net for decomposition and an Enhance-Net for illumination adjustment. Guo \textit{et al.} \cite{dark_camera0} proposes Zero-DCE which formulates light enhancement as a task of image-specific curve estimation with a deep network. Retinexformer \cite{Cai2023LLIE} formulates a simple yet principled One-stage Retinex-based Framework to light up low-light images. Besides, some work focuses on the robustness of vision tasks to low-light, \eg object detection. Wang \textit{et al.} \cite{wang2023research} combines with the image enhancement algorithm to improve the accuracy of object detection.  For \textbf{spike camera}, it is also affected by low-light environments. Dong \textit{et al.} \cite{lowlighrec0} propose a real low-light high-speed dataset for reconstruction. However, it lacks corresponding image sequences as ground truth. Besides, the concurrent work \cite{rec_mm} synthesizes a low-light spike stream dataset. However, it only contains static scenes and cannot evaluate the performance of reconstruction methods during motion. 
%
\subsection{Reconstruction for Spike Camera}
The reconstruction of high-speed dynamic scenes has been a popular topic for spike camera. Based on the statistical characteristics of spike stream, Zhu \textit{et al.} \cite{spikecamera} first reconstruct high-speed scenes. Zhao \textit{et al.} \cite{rec0} improved the smoothness of reconstructed scenes by introducing motion aligned filter. Zhu \textit{et al.} \cite{rec1} construct a dynamic neuron extraction model to distinguish the dynamic and static scenes. With the rise of spiking neural networks \cite{xu2024snn,shen2023snn,shen2024snn,zhu2024esnn}, for enhancing reconstruction results, Zheng \textit{et al.} \cite{rec2} uses short-term plasticity mechanism to exact motion area. Zhao \textit{et al.} \cite{rec3} first proposes a deep learning-based reconstruction framework, Spk2ImgNet (S2I), to handle the challenges brought by both noise and high-speed motion. Chen \textit{et al.} \cite{rec5} build a self-supervised reconstruction framework by introducing blind-spot networks. It achieves desirable results compared with S2I. The reconstruction method \cite{rec6} presents a novel Wavelet Guided Spike Enhancing (WGSE) paradigm. By using multi-level wavelet transform, the noise in the reconstructed results can be effectively suppressed. Besides, we would like to mention the concurrent work, RSIR \cite{rec_mm}. In RSIR, the AST representation is used to adaptively extract the number of spike in a spike stream under different illuminations. Then, a multi-scale wavelet recurrent network can reconstruct images from the AST representation. However, 
AST compresses a spike stream into a spike number map which ignores dynamic information, resulting in more motion blur for high-speed low-light scenes. This greatly limits the contribution of RSIR to spike camera reconstruction, as the original intention of spike camera is to handle high-speed dynamic scenes. Unlike AST, our proposed representation, LR-Rep, first calculates global inter-spike interval map (GISI). It can better preserve dynamic information while aggregating temporal information.  




%

\subsection{Spike Camera Simulation}
Spike camera simulation is a popular way to generate spike streams and accurate labels. Zhao \textit{et al.} \cite{rec3} first convert interpolated image sequences with high frame rate into spike stream. Based on \cite{rec3}, the simulators \cite{rec4,2021spk_sim,zhao2022spikingsim} add some random noise to generate spike streams more accurately. To avoid motion artifacts caused by interpolation, Hu \textit{et al.} \cite{2022scflow} presents the spike camera simulator (SPCS) combining simulation function and rendering engine tightly. Then, based on SPCS, optical flow datasets for spike camera are first proposed. Zhang \textit{et al.} \cite{2022spkTransformer} generate the first spike-based depth dataset by the spike camera simulation. Zhang \textit{et al.} \cite{rec6} generate the first semantic segmentation spike streams dataset by the spike camera simulation.

\section{Reconstruction Datasets}
In order to train and evaluate the performance of reconstruction methods in low-light high-speed scenes, we propose two low-light spike stream datasets, \textbf{R}and \textbf{L}ow-\textbf{L}ight \textbf{R}econstruction (RLLR) and \textbf{L}ow-\textbf{L}ight \textbf{R}econstruction (LLR) based on spike camera model. RLLR is used as our train dataset and LLR is carefully designed to evaluate the performance of different reconstruction methods as test dataset. 
We first introduce the spike camera model, and then introduce our datasets where noisy spike streams are generated by the spike camera model.


\hupar{Spike camera model} Each pixel on the spike camera model converts light signal into the current signal and accumulates the input current. For pixel  $\mathbf{x} = (x, y)$, if the accumulation of input current reaches a fixed threshold  $\phi$, a spike is fired and then the accumulation can be reset as,
   \begin{align}
    &{A}(\mathbf{x}, t) = {A_\mathbf{x}}(t) \; {\rm{mod}} \; \phi = \int_{0}^{t} {I_{tot}}(\mathbf{x}, \tau) d\tau\ {\rm{mod}} \; \phi, 
    \\
    &{I_{tot}}(\mathbf{x}, \tau) = {I_{in}}(\mathbf{x}, \tau) + {I_{dark}}(\mathbf{x}, \tau),
    \end{align}
where ${A}(\mathbf{x}, t)$ is the accumulation at time $t$, ${A_\mathbf{x}}(t)$ is the accumulation without reset before time $t$, ${I_{in}}(\mathbf{x}, \tau)$ is the input current at time $\tau$ (proportional to light intensity) and ${I_{dark}}(\mathbf{x}, \tau)$ is the main fixed pattern noise in spike camera, \ie dark current \cite{rec4, zhao2022spikingsim, hu2024scsim}. Further, due to limitations of circuits, each spike is read out at discrete time $nT, n \in \mathbb{N}$ ($T$ is a micro-second level). Thus, the output of the spike camera is a spatial-temporal binary stream $S$ with $H \times W \times N$ size. The $H$ and $W$ are the height and width of the sensor, respectively, and $N$ is the temporal window size of the spike stream. According to the spike camera model, it is natural that the spikes (or information) in low-light spike streams are sparse because reaching the threshold is lengthy. 

\begin{figure*}[!t]
\includegraphics[width=\linewidth]{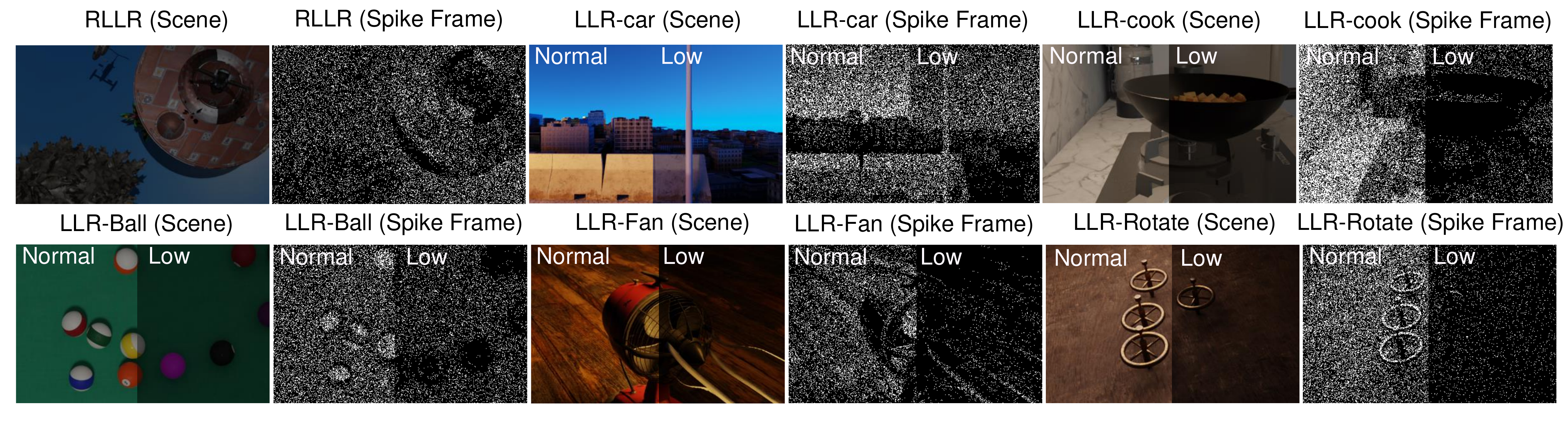}
\centering
\caption{Proposed datasets, RLLR and LLR. RLLR includes random scenes and LLR includes carefully designed scenes. A Spike Frame is a slice of generated spike streams on a temporal axis.}\label{dataset}
\end{figure*}

\hupar{RLLR}
As shown in Fig.~\ref{dataset}, RLLR includes 100 random low-light high-speed scenes where high-speed scenes are first generated by SPCS \cite{2022scflow} and then the light intensity of all pixels in each scene is darkened by multiplying a random constant ($0$-$1$).  Each scene in RLLR continuously records a spike stream with $400 \times 250 \times 1000$ size and corresponding image sequence. Then, for each image, we clip a spike stream with $400 \times 250 \times 41$ size from the spike stream as input.

\hupar{LLR}
As shown in Fig.~\ref{dataset}, LLR includes 5$\times$2 carefully designed high-speed scenes where we use the scenes with five kinds of motion (named Ball, Car, Cook, Fan, and Rotate) and each scene corresponds to two light sources (normal and low). To ensure the reliability of our scenes, different light sources are used, and the power of the light source is consistent with the real world. Besides, the motion in Ball, Cook, Fan, and Rotate is from \cite{2022scflow} while the motion in Car is created based on vehicle speed in the real world. Hence, the motion of objects is close to the real world. Each scene in LLR continuously records 21 spike streams with $400 \times 250 \times 41$ size and 21 corresponding images. In the proposed datasets, we consider the noise of spike camera based on \cite{zhao2022spikingsim}. 

\section{Method}
\begin{figure*}[htbp]
\includegraphics[width=\linewidth]{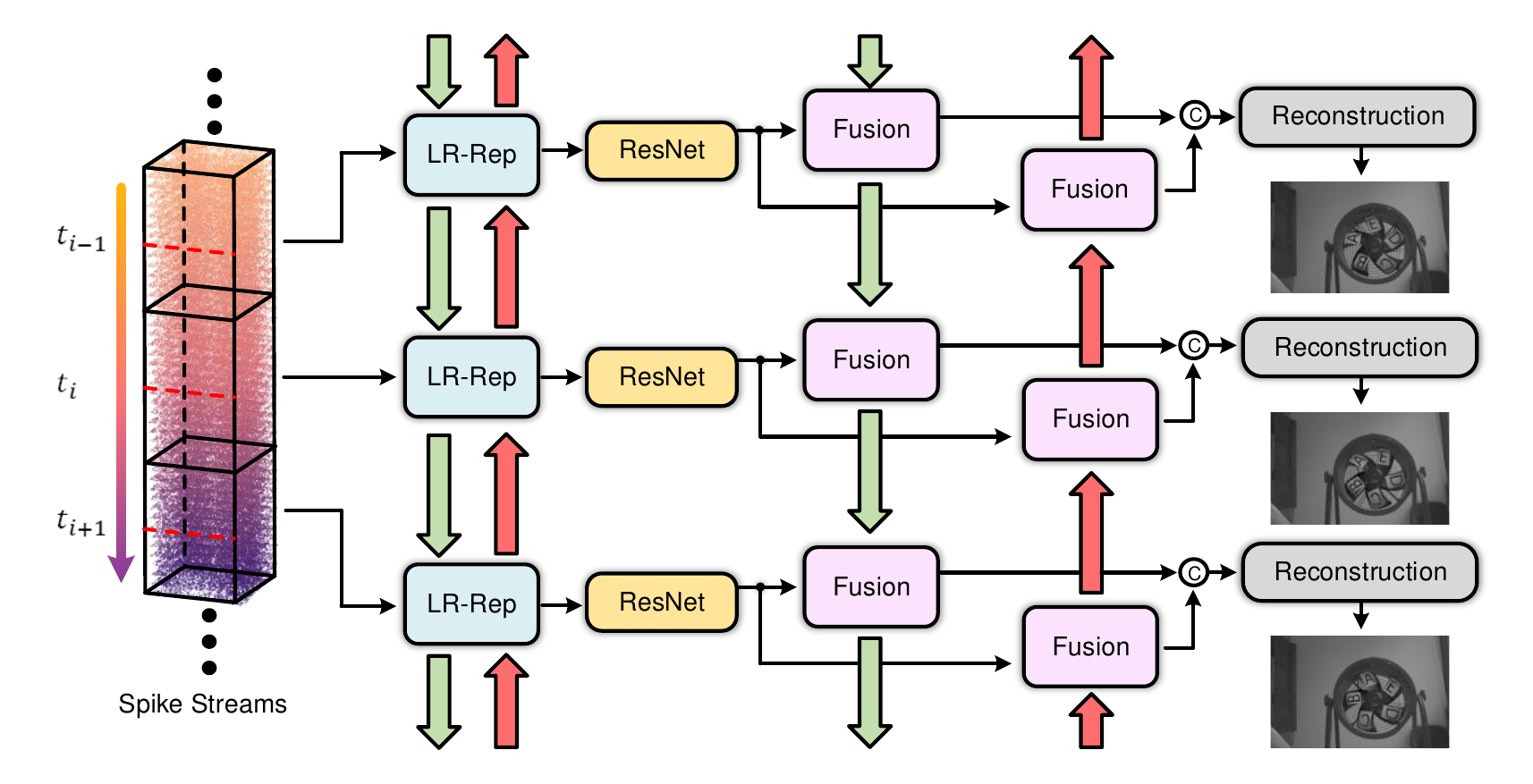}
\centering
\caption{{Illustration of the proposed bidirectional recurrent-based reconstruction framework. It includes a light-robust representation, feature extractor (ResNet), fusion, and reconstruction. The green and red lines represent the forward and backward data flow. The two kinds of data flow are independent.}}\label{rr_new}
\end{figure*}

\subsection{Problem Statement} 

For simplicity, we write $\mathbf{S}_{t} \in \{0, 1\}^{H \times W \times (2\Delta t + 1)}$ to denote a spike stream from time $t - \Delta t$ to $t + \Delta t$ ($2\Delta t + 1$ is the fixed temporal window) and write $\mathbf{Y}_t \in \mathbb{R}^{H \times W}$ to denote the instantaneous light intensity received in spike {{camera}} at time $t$. Reconstruction is to use continuous spike streams, $\{\mathbf{S}_{t_i}, t_i = i * (2\Delta t + 1) | i =1,2,3...K\}$ to restore the light intensity information at different time, $\{\mathbf{Y}_{t_i}, t_i = i * (2\Delta t + 1) | i =1,2,3...K\}$. Generally, the temporal window $2\Delta t + 1$ is set as 41 which is the same with \cite{rec3,rec5,rec6}.

\subsection{Overview}
To overcome the challenge of low-light spike streams, \ie the recorded information is sparse (see Fig.\ref{teasor}), we propose a light-robust reconstruction method that can fully utilize temporal information of spike streams. It is beneficial from two modules: 1. A light-robust representation, LR-Rep. 2. A fusion module. As shown in Fig.~\ref{rr_new}, to recover the light intensity information at time $t_i$, $\mathbf{Y}_{t_i}$, we first calculate the light-robust representation at time $t_i$, written as $\mathbf{Rep}_{t_i}$. Then, we use a ResNet module to extract deep features, $\mathbf{F}_{t_i}$, from $\mathbf{Rep}_{t_i}$. $\mathbf{F}_{t_i}$ is fused with forward (backward) temporal features as $\mathbf{F}_{t_i}^f$ ($\mathbf{F}_{t_i}^b$). Finally, we reconstruct the image at time ${t_i}$, $\mathbf{\hat{Y}}_{t_i}$ with $\mathbf{F}_{t_i}^f$ and $\mathbf{F}_{t_i}^b$.




\begin{figure*}[t]
\includegraphics[width=0.9\linewidth]{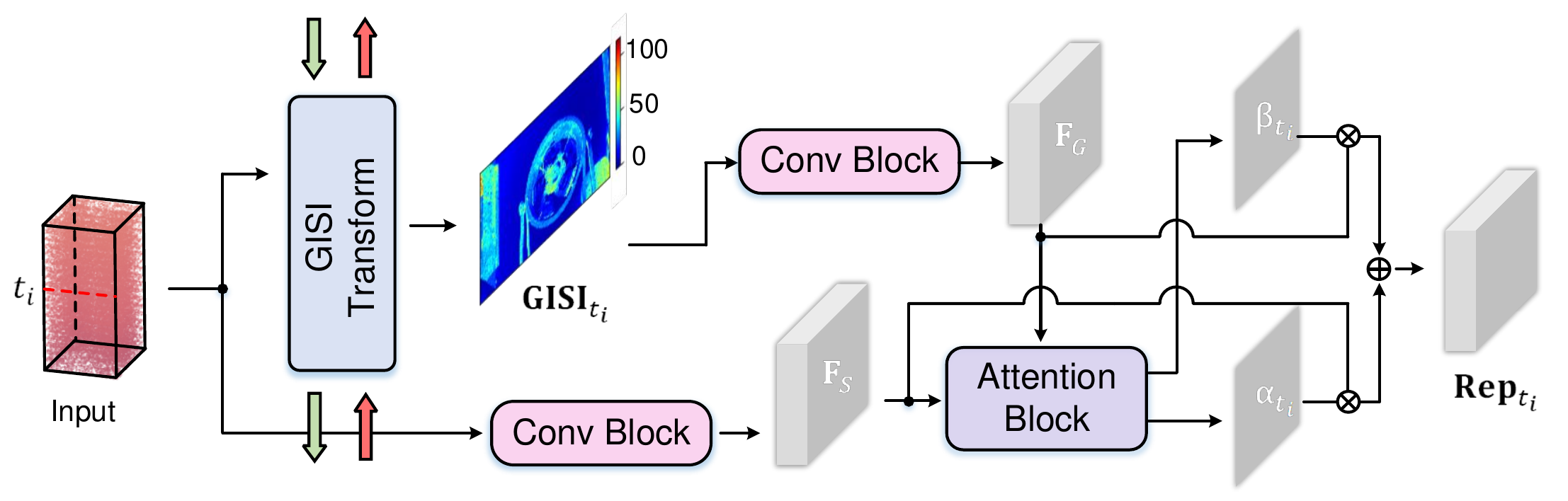}
\centering
\caption{{Illustration of the proposed light-robust representation. We use convolution blocks to extract shallow features from input spike stream and GISI, respectively. Then they are fused by an attention block.}}\label{rep}
\end{figure*}

\begin{figure*}[htbp]
\includegraphics[width=1.0\linewidth]{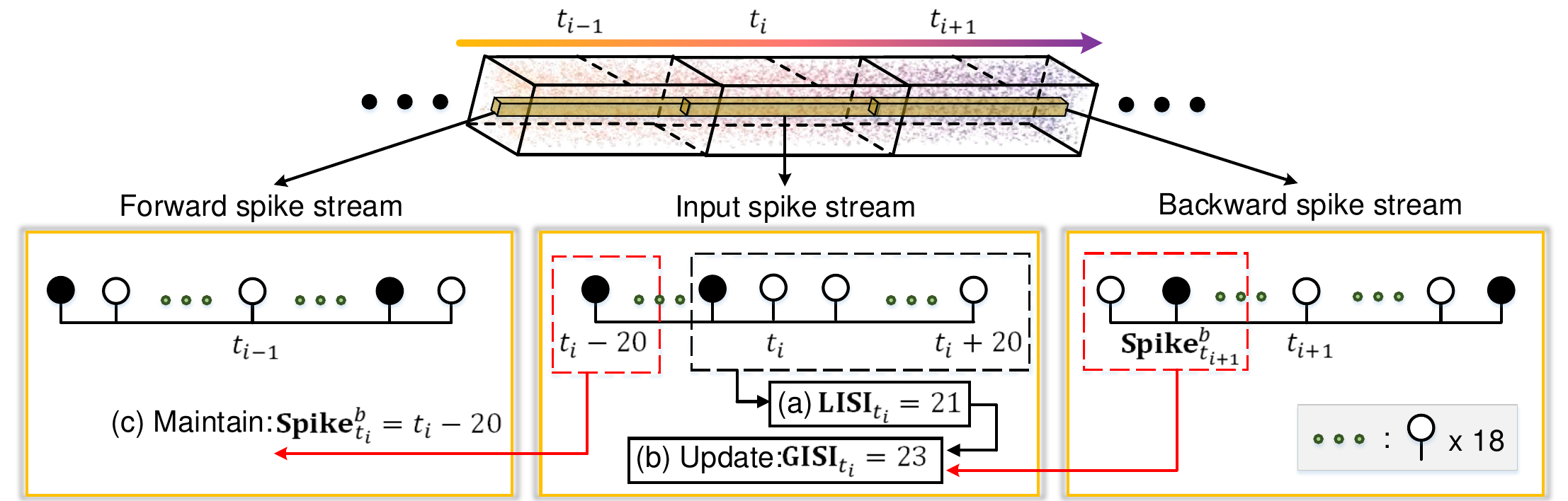}
\centering

\caption{{Illustration of GISI transform for backward in a pixel. (a). Calculate the local inter-spike interval, $\mathbf{LISI}_{t_i}$ from the input spike stream \cite{rec5,zhao2022learning}. (b). Update global inter-spike interval, $\mathbf{GISI}_{t_i}$ based on the release time of backward spike, $\mathbf{Spike}_{t_{i+1}}^b$ and $\mathbf{LISI}_{t_i}$. (c). Maintain and transmit the release time of backward spike, $\mathbf{Spike}_{t_{i}}^b$. Black (white) circle is a (no) spike and the red line is backward data flow.}}\label{gisi}

\end{figure*}

\subsection{Light-robust Representation}

\begin{figure*}[t]
\centering
\includegraphics[width=0.9\linewidth]{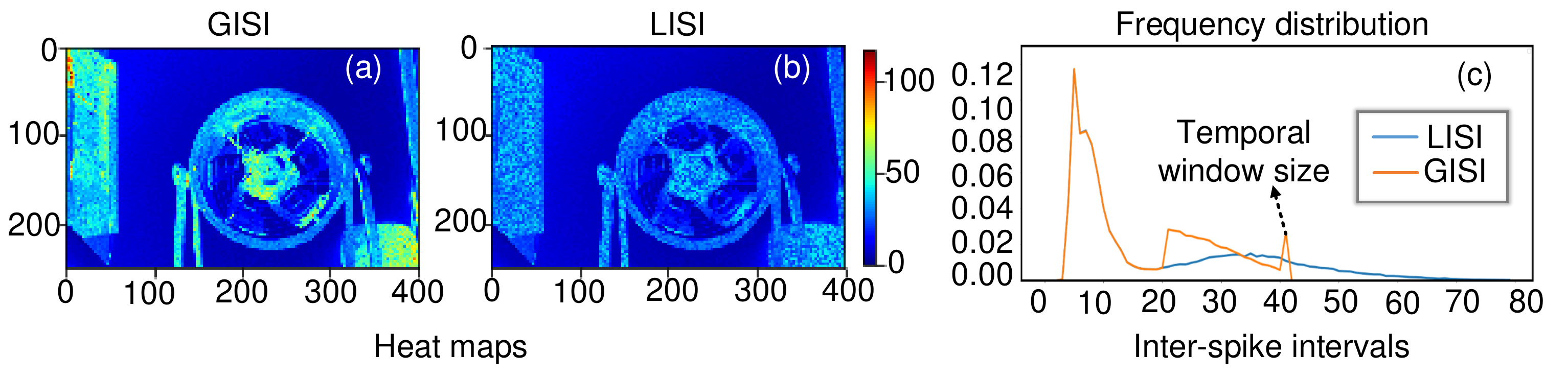}
\caption{(a) and (b) show the visualizations of $\mathbf{GISI}_{t_i}$ and $\mathbf{LISI}_{t_i}$ in a real spike stream. (c) shows the distribution of pixel-wise values in $\mathbf{GISI}_{t_i}$ and $\mathbf{LISI}_{t_i}$.}\label{gisi_lisi}
\end{figure*}

As shown in Fig.~\ref{rep}, a light-robust representation, LR-Rep, is proposed to aggregate the information in low-light spike streams. 
{{LR-Rep mainly consists of two parts, GISI transform and feature extraction.

\hupar{GISI transform} Calculating the local inter-spike interval from the input spike stream is a common operation \cite{rec5,zhao2022learning} and we call it as LISI transform. Different from LISI transform, we propose a GISI transform that can utilize the release time of forward and backward spikes to obtain the global inter-spike interval $ \mathbf{GISI}_{t_i}$. It needs to be performed twice, i.e. once forward and once backward respectively. Taking GISI transform backward as an example, it can be summarized as three steps as shown in Fig.~\ref{gisi}. GISI transform can extract more temporal information from spike streams than LISI transform as shown in Fig.~\ref{gisi_lisi}. 


%

\hupar{Feature extraction}}} After GISI transform, we separately extract shallow features of $\mathbf{GISI}_{t_i}$ and input spike stream, $\mathbf{F}_G$ and $\mathbf{F}_S$ through convolution block. Finally, $\mathbf{Rep}_{t_i}$ is obtained by an attention module where  $\mathbf{F}_G$ and $\mathbf{F}_S$ are integrated, \ie
\begin{align}
&[\mathbf{ \upbeta}_{t_i}, \mathbf{\upalpha}_{t_i}] = Att([\mathbf{F}_G, \mathbf{F}_S]),\\
&\mathbf{Rep}_{t_i} = \mathbf{\upbeta}_{t_i}\mathbf{F}_G + \mathbf{\upalpha}_{t_i}\mathbf{F}_S,
\end{align}
where $Att(\cdot)$ denotes an attention block including 3-layer convolution with 3-layer activation function and $\mathbf{Rep}_{t_i}$ is our LR-Rep at time ${t_i}$.

\subsection{Fusion and Reconstruction}

\begin{figure}[htb]
\includegraphics[width=0.95\linewidth]{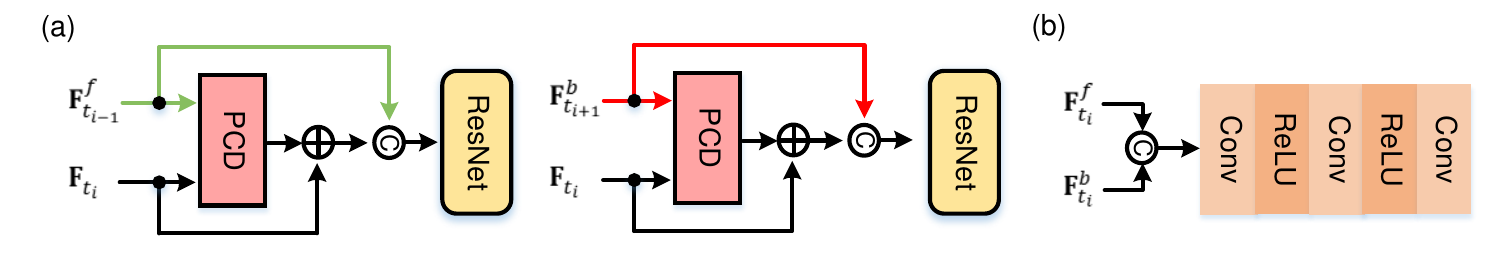}
\centering
\caption{{Illustration of fusion module and reconstruction module. (a) denotes forward (green line) and backward (red line) fusion modules. (b) denotes the reconstruction module.}}\label{fusion}
\end{figure}

We first extract the deep feature $\mathbf{F}_{t_i}$ of $\mathbf{Rep}_{t_i}$ through a ResNet with 16 layers. Then, as shown in Fig.~\ref{fusion}(a), for forward, temporal features $\mathbf{F}_{t_{i-1}}^f$ and $\mathbf{F}_{t_i}$ are fused as temporal features of the input spike stream $\mathbf{F}_{t_i}^f$. For backward, temporal features $\mathbf{F}_{t_{i+1}}^b$ and $\mathbf{F}_{t_i}$ are fused as temporal features of the input spike stream $\mathbf{F}_{t_i}^b$. To avoid the misalignment of motion from different timestamps, we use a Pyramid Cascading and Deformable convolution (PCD) \cite{Wang_2019_CVPR_Workshops} to add alignment information to $\mathbf{F}_{t_i}$. The above process can be written as,
\begin{align}
&\mathbf{F}_{t_i} = f(\mathbf{Rep}_{t_i}),\\
&\mathbf{F}_{t_i}^f = f([\mathbf{F}_{t_i} + a(\mathbf{F}_{t_{i-1}}^f, \mathbf{F}_{t_i}), \mathbf{F}_{t_{i-1}}^f]),\\
&\mathbf{F}_{t_i}^b = f([\mathbf{F}_{t_i} + a(\mathbf{F}_{t_{i+1}}^b, \mathbf{F}_{t_i}), \mathbf{F}_{t_{i+1}}^b]),
\end{align}
where $f(\cdot)$ denotes the feature extraction and $a(\cdot, \cdot)$ denotes the PCD module. Finally, as shown in Fig.~\ref{fusion}(b), we use forward and backward temporal features ($\mathbf{F}_{t_i}^b$ and $\mathbf{F}_{t_i}^f$) to reconstruct the current scene at time $t_i$, \ie
\begin{align}
&\mathbf{\hat{Y}}_{t_i} = c([\mathbf{F}_{t_i}^b, \mathbf{F}_{t_i}^f]),\\
& \mathcal{L} = \sum\limits_{i=1}^{K} \Vert \mathbf{\hat{Y}}_{t_i} - \mathbf{{Y}}_{t_i} \Vert_1
\end{align}
where $c(\cdot)$ denotes 3-layer convolution with 2-layer
ReLU, $\mathcal{L}$ is the loss function,  $\Vert \cdot \Vert_1$ denotes 1-norm and $K$ is the number of continuous spike streams.

\section{Experiment}

\subsection{Implementation Details}
We train our method in the proposed dataset, RLLR. Consistent with previous work \cite{rec3, rec5,rec6}, the temporal window of each input spike stream is 41. The spatial resolution of input spike streams is randomly cropped the spike stream to $64 \times 64$ during the training procedure and the batch size is set as 8. Besides, forward (backward) temporal features and the release time of spikes in our method are maintained from 21 continuous spike streams. We use Adam optimizer with $\beta_1 = 0.9$ and $\beta_2 = 0.99$. The learning rate is initially set as 1e-4 and scaled by 0.1 after 70 epochs. The model is trained for 100 epochs on 1 NVIDIA A100-SXM4-80GB GPU.

\subsection{Results}



We compare our method with {traditional} reconstruction methods, \ie TFI \cite{spikecamera}, STP \cite{rec2}, SNM \cite{rec1} and {deep learning-based} reconstruction methods, \ie SSML \cite{rec5}, Spk2ImgNet (S2I) \cite{rec3}, WGSE \cite{rec6}, concurrent work RSIR \cite{rec_mm}. The supervised learning methods, S2I, WGSE and RSIR, are trained on RLLR. We evaluate methods on two kinds of data: 
\\{(1)} The carefully designed \textbf{synthetic} dataset, LLR.
\\{(2)} The \textbf{real} spike streams dataset, PKU-Spike-High-Speed \cite{rec3} and low-light high-speed spike streams dataset \cite{lowlighrec0}.  

The \textbf{reproduction} of these methods is from their official source codes.

  
  

\begin{table}[htbp]
  \centering
  \caption{{PSNR and SSIM of reconstruction results on synthetic dataset, LLR. The best performance is bolded. Note that high PSNR (S2I, WGSE, and Ours are above 40) is a normal occurrence due to low-light scenes (see appendix).}}
  \resizebox{0.9\columnwidth}{!}{
    \begin{tabular}{ccccccccc}
    \toprule
    \multirow{2}[0]{*}{\textbf{Metric}}  & \textbf{TFI}   & \textbf{RSIR}  & \textbf{SSML}  & \textbf{S2I}   & \textbf{STP}   & \textbf{SNM}   & \textbf{WGSE}  & \textbf{Ours} \\
          & \textbf{ICME$^{,}$19} & \textbf{MM$^{,}$23} & \textbf{IJCAI$^{,}$22} & \textbf{CVPR$^{,}$21} & \textbf{CVPR$^{,}$21} & \textbf{PAMI$^{,}$22} & \textbf{AAAI$^{,}$23} & \textbf{This paper} \\
    \midrule
    PSNR  & 31.409 & 34.121 & 38.432 & 40.883 & 24.882 & 25.741 & 42.959 & \textbf{45.075} \\
    SSIM  & 0.72312 & 0.88337 & 0.89942 & 0.95915 & 0.55537 & 0.80281 & 0.97066 & \textbf{0.98681} \\
    \bottomrule
    \end{tabular}%
   }
  \label{metric}%
\end{table}%

\begin{figure}[tbp]
\includegraphics[width=1.0\linewidth]{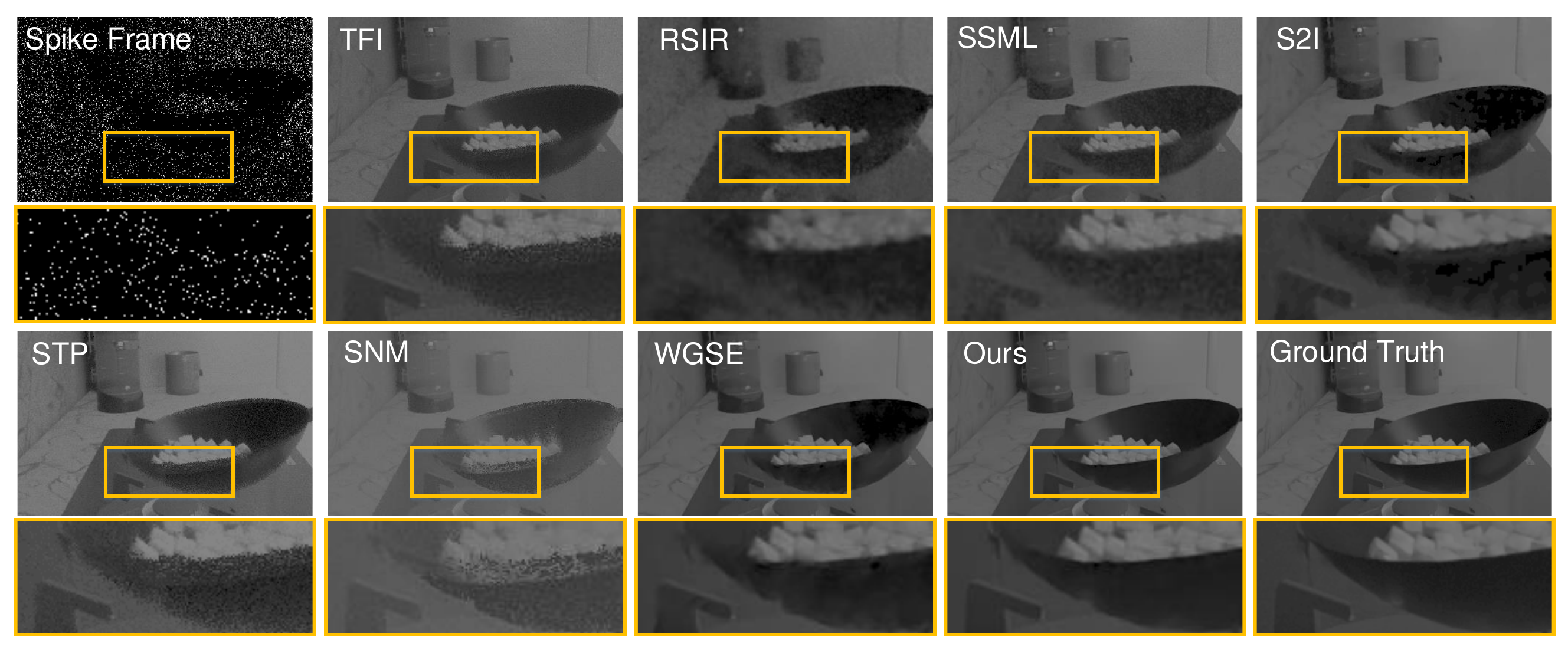}
\centering
\caption{An example of different methods on the LLR. Spike Frame is a slice of input spike streams on
temporal axis. Please enlarge for details. More results are in appendix.}\label{results}
\end{figure}

\hupar{Results on our synthetic dataset} As shown in Table.~\ref{metric}, we use the two reference image quality assessment (IQA) metrics, \ie PSNR and SSIM to evaluate the performance of different methods on LLR. We can find that our method achieves the best reconstruction performance and has a PSNR gain over 2dB than the state-of-the-art reconstruction method, WGSE, which demonstrates its effectiveness. Fig.~\ref{results} shows the visualization results from different reconstruction methods. We can find that our method can better restore motion details in low-light motion regions than other methods. Besides, RSIR is designed to handle spike streams in static scenes and we find that it can suffer from large motion blur in low-light high-speed scenes.



\begin{figure}[!ht]
\includegraphics[width=\linewidth]{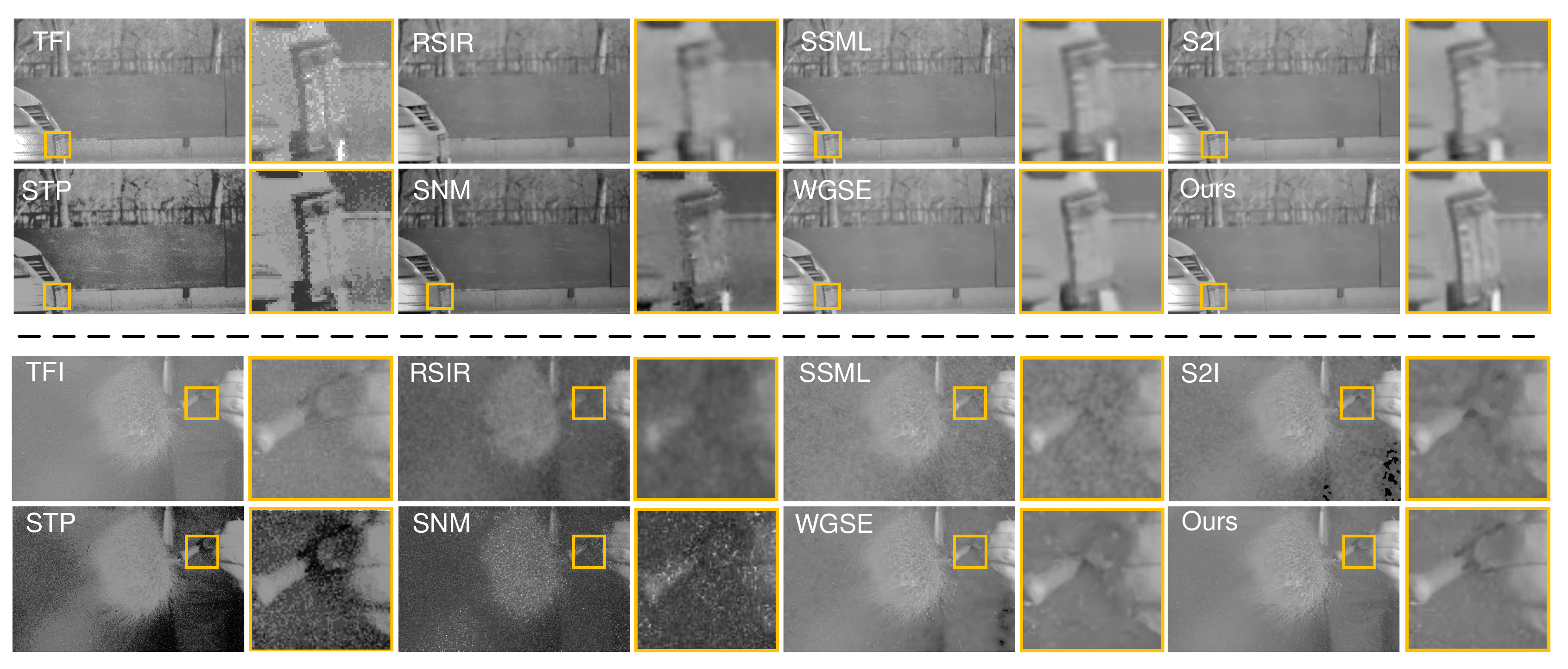}
\centering
\caption{Results from different reconstruction methods on the real datasets, PKU-Spike-High-Speed \cite{rec3} (Top) and low-light high-speed spike streams dataset \cite{lowlighrec0} (Bottom). For low-light high-speed spike streams dataset, we apply the traditional enhancement method \cite{ying2017hdr} to reconstruction results because the scene is too dark. More results are in our appendix.}\label{results_real_pku}
\end{figure}

\hupar{Results on real datasets}
For real data, we test different methods on two spike stream datasets, PKU-Spike-High-Speed \cite{rec3} and low-light spike streams \cite{lowlighrec0}. PKU-Spike-High-Speed includes 4 high-speed scenes under normal-light conditions and \cite{lowlighrec0} includes 5 high-speed scenes under low-light conditions. Fig.~\ref{results_real_pku} shows the reconstruction results. Note that we apply the traditional enhancement method \cite{ying2017hdr} to reconstruction results on \cite{lowlighrec0} because scenes are too dark. We can find that, for high-speed scenes under normal-light conditions, deep learning-based methods (SSML, RSIR, S2I, WGSE, and Ours) can reconstruct scene details well. However, for high-speed scenes under low-light conditions, SSML and RSIR introduce a large amount of motion blur while S2I and WGSE may introduce some artifacts in dark backgrounds. Our method can more effectively restore the information in scenes, i.e., clear texture.

\begin{figure}[h]
\includegraphics[width=0.75\linewidth]{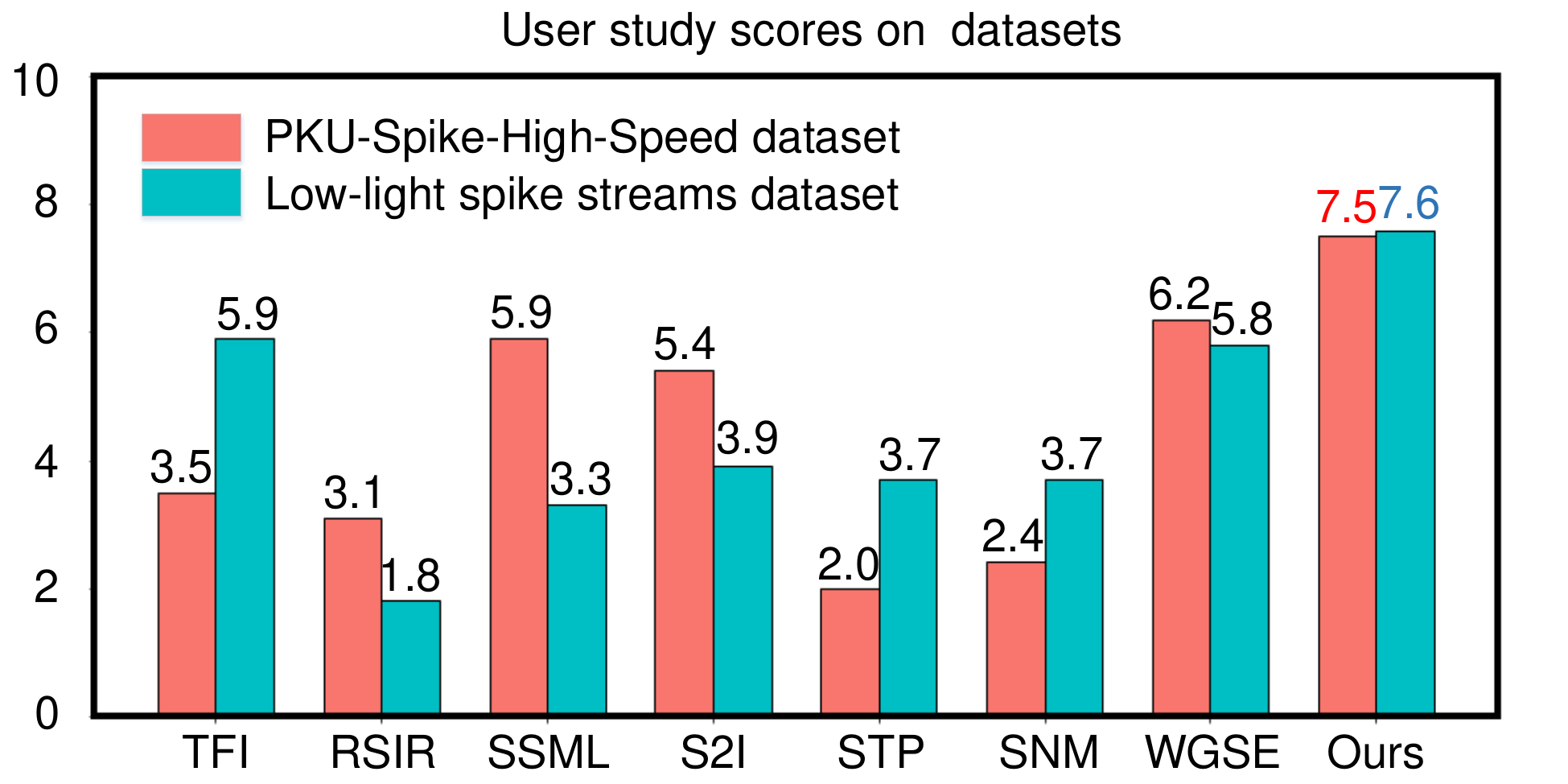}
\centering
\caption{User study scores ($\uparrow$) of reconstructed images from different methods. The max (min) score is 8 (1). Red or blue color is the highest score on the dataset \cite{rec3} or \cite{lowlighrec0}.}\label{US}
\end{figure}

As shown in Fig.~\ref{US}, we perform a user study written as US \cite{wilson1981user, jiang2021enlightengan} to quantify the visual quality of different methods. For each scene in datasets, we randomly select reconstructed images at the same time from different methods and display them on the screen (the image order is randomly shuffled). 20 human subjects (university degree or above) are invited to independently score the visual quality of the reconstructed image. The scores of visual quality range from 1 to 8 (worst to best quality). The average subjective scores for each spike stream dataset are shown in Fig.~\ref{US} and our method reaches the highest US score in all methods.

\begin{figure*}[!h]
\includegraphics[width=0.8\linewidth]{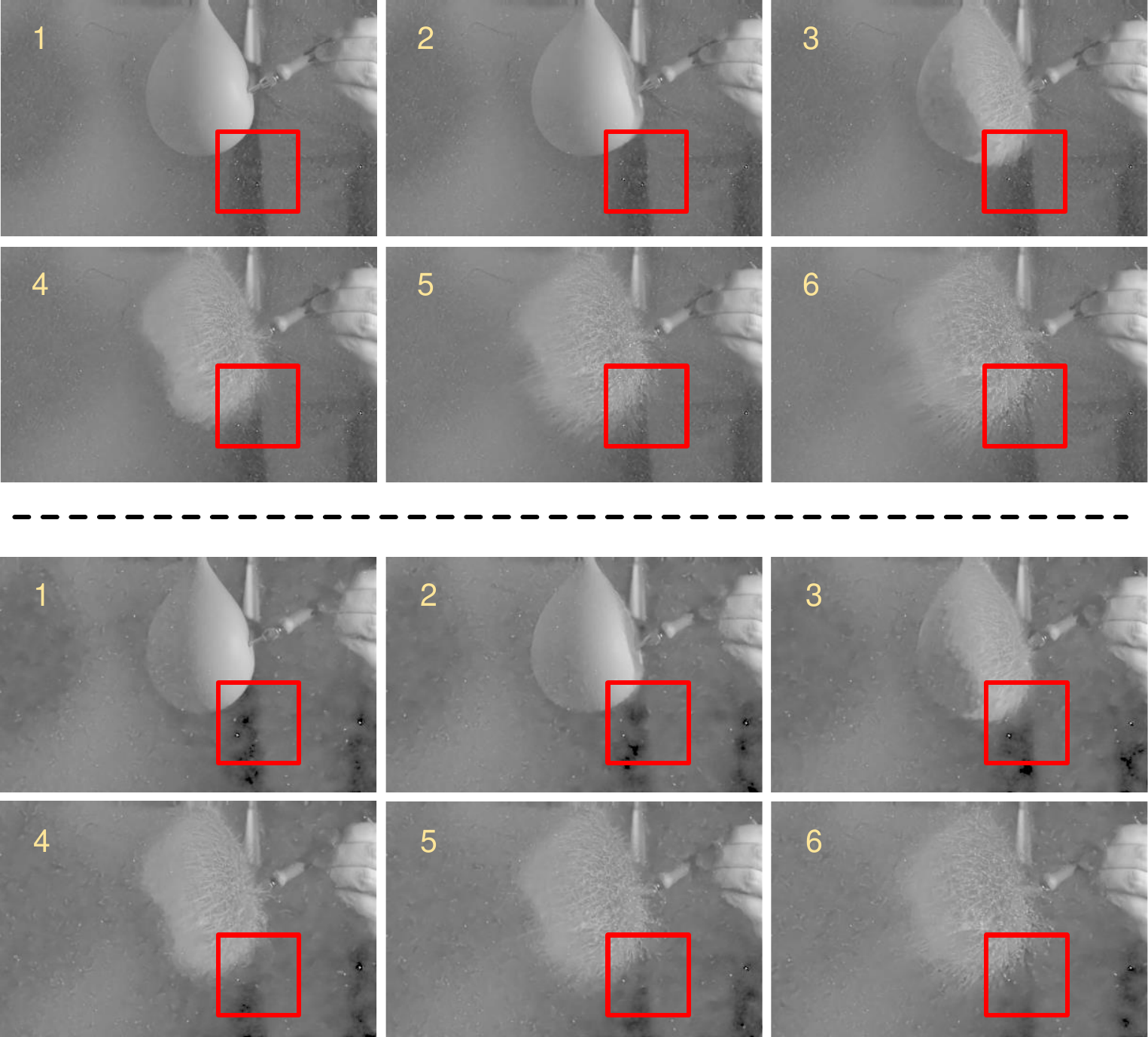}
\centering
\caption{A water polo bursting at high speed in a low-light indoor. We selected the reconstruction results under 6 sampling moments, and the interval between two adjacent sampling moments is 41/40000 s. The \textbf{top} is our method and the \textbf{bottom} is the state-of-the-art reconstruction method \cite{rec6}. We apply the traditional enhancement method \cite{ying2017hdr} to reconstruction results because the scene is too dark. Reconstructed videos are provided in supplementary materials.}\label{result_video}
\end{figure*}

\hupar{Temporal consistency of reconstructed results} Our reconstruction method is stable to spike stream at different moments. Fig.~\ref{result_video} shows the continuous reconstructed results in a real high-speed low-light scene. We find that our method can recover scene details at different moments, while the state-of-the-art WGSE introduces temporal-varying artifacts. Besides, we also provide a reconstruction video in our supplementary material.



\subsection{Ablation}

    \hupar{Proposed modules} To investigate the effect of the proposed light-robust representation LR-Rep, the \textbf{a}djacent (forward and backward) \textbf{d}eep temporal \textbf{f}eatures (ADF), \ie  $\mathbf{F}_{t_i}^b$ and $\mathbf{F}_{t_i}^f$ in our fusion module, the \textbf{a}lignment \textbf{i}nformation in our \textbf{f}usion module (AIF) and GISI transform in LR-Rep, we compare 5 baseline methods with our final method. (A) is the basic baseline without LR-Rep, ADF, and AIF. Table.~\ref{ablation} shows ablation results on the proposed dataset, LLR. The comparison between (A) and (C) ((B) and (D)) proves the effectiveness of LR-Rep. The comparison between (A) and (B) ((C) and (D)) proves the effectiveness of ADF. Further, by adding the alignment information in the fusion module \ie AIF, our final method (E) appropriately reduces the misalignment of motion from different timestamps and can reconstruct high-speed scenes more accurately than (D). Besides, the comparison between (E) and (F) shows GISI has better performance than LISI. This is because GISI can extract more temporal information than LISI (see Fig.~\ref{gisi_lisi}). More importantly, the cost of using GISI instead of LISI is negligible (we only need to use two 400$\times$250 matrices to store the time of the forward spike and the backward spike, respectively), which does not affect the parameter and efficiency of the network. 

\begin{table}[htbp]
  \centering
  \caption{Abltion results on the synthetic dataset, LLR. The best performance is bolded.}
  \resizebox{0.7\columnwidth}{!}{
    \begin{tabular}{llcc}
    \toprule
    \multicolumn{1}{c}{\textbf{Index}} & \textbf{Effect of different network structures} & \textbf{PSNR}  & \textbf{SSIM} \\
    \midrule
    (A)   &  Basic baseline & 42.743
 & 0.97403
 \\
    (B)   &  Adding ADF  to (A)  & 44.151 & 0.98514 \\
    (C)   &   Adding LR-Rep to (A)  & 44.739

 & 0.98636

 \\
    (D)   &  Adding ADF \& LR-Rep to (A) & 44.956
 & 0.98678 \\
    (E)   &  Adding ADF \& LR-Rep \& AIF  & \bf 45.075 & \bf 0.98681
 \\
    (F)   & Replacing GISI with LISI in (E) & 44.997
 & 0.98676 \\
 \bottomrule
    \end{tabular}%
  }
  \label{ablation}%
\end{table}%

\hupar{Comparison with other representation}  We compare the performance of different representations in our framework, \ie (1) General representation of spike stream: TFI and TFP \cite{spikecamera} (2) Tailored representation for reconstruction networks: AST in RSIR \cite{rec_mm}, AMIM \cite{rec5} in SSML, SALI \cite{rec3} in S2I and WGSE-1d \cite{rec6} in WGSE. We replace LR-Rep in our method as the above representation. They are trained on the dataset, RLLR, and implementation details are the same as our method.  As shown in Table.~\ref{rep_ab}, our LR-Rep achieves the best performance which means LR-Rep can better adapt to our framework.

\begin{table}[ht]
  \centering
  \caption{Performance of different representation methods in our framework. All methods are trained on RLLR and are tested on LLR. The best performance is bolded.}
  \resizebox{0.85\columnwidth}{!}{
    \begin{tabular}{l|ccccccc}
    \toprule
        \multirow{2}[0]{*}{\textbf{Rep.}}  & \textbf{TFP}   & \textbf{TFI}  & \textbf{AST}  & \textbf{AMIM}   & \textbf{SALI}   & \textbf{WGSE-1d}   & \textbf{LR-Rep} \\
          & \textbf{ICME$^{,}$19} & \textbf{ICME$^{,}$19} & \textbf{MM$^{,}$23} & \textbf{IJCAI$^{,}$22}  & \textbf{CVPR$^{,}$21} & \textbf{AAAI$^{,}$23} & \textbf{Ours} \\
    \midrule
    PSNR  & 38.615 & 37.617 & 37.997 & 41.950 & 43.314 & 42.302 & \bf 45.075 \\
    SSIM  & 0.96641 & 0.93632 & 0.95463 & 0.97493 & 0.98304 & 0.97438 & \bf 0.98681 \\
    \bottomrule
    \end{tabular}%
   }
  \label{rep_ab}%
\end{table}%

\begin{table}[htbp]
  \centering
  \caption{Evaluation results on LLR. We train our network where 20\%, 40\%, 60\%, and 80\% of RLLR data are used as training set respectively. The best performance is bolded.}
  \resizebox{0.55\columnwidth}{!}{
    \begin{tabular}{cccccc}
    \toprule
    \textbf{Metric} & \textbf{20}\%  & \textbf{40}\%  & \textbf{60}\%  & \textbf{80}\%  & \textbf{100}\% \\
    \midrule
    PSNR  & 35.001 & 38.618 & 44.415 & 44.753 & \textbf{45.075} \\
    SSIM  & 0.93411 & 0.97113 & 0.98459 & 0.98581 & \textbf{0.98681} \\
    \bottomrule
    \end{tabular}%
   }
  \label{datasetsize}%
\end{table}%


\hupar{Train dataset size.} The size of train datasets has an impact on the performance of our network. A larger train dataset typically provides more samples and a wider range of variations. In fact, proposed RLLR is enough for the reconstruction task of low-light spike streams.  As shown in Table.~\ref{datasetsize}, we find that as the dataset size increase, the performance of the model also improves. However, it is observed that the performance improvement becomes less significant after the dataset size reaches 60\% of RLLR. It shows that the proposed RLLR is sufficient for training our network.


\begin{figure}[tbp]
\includegraphics[width=1.0\linewidth]{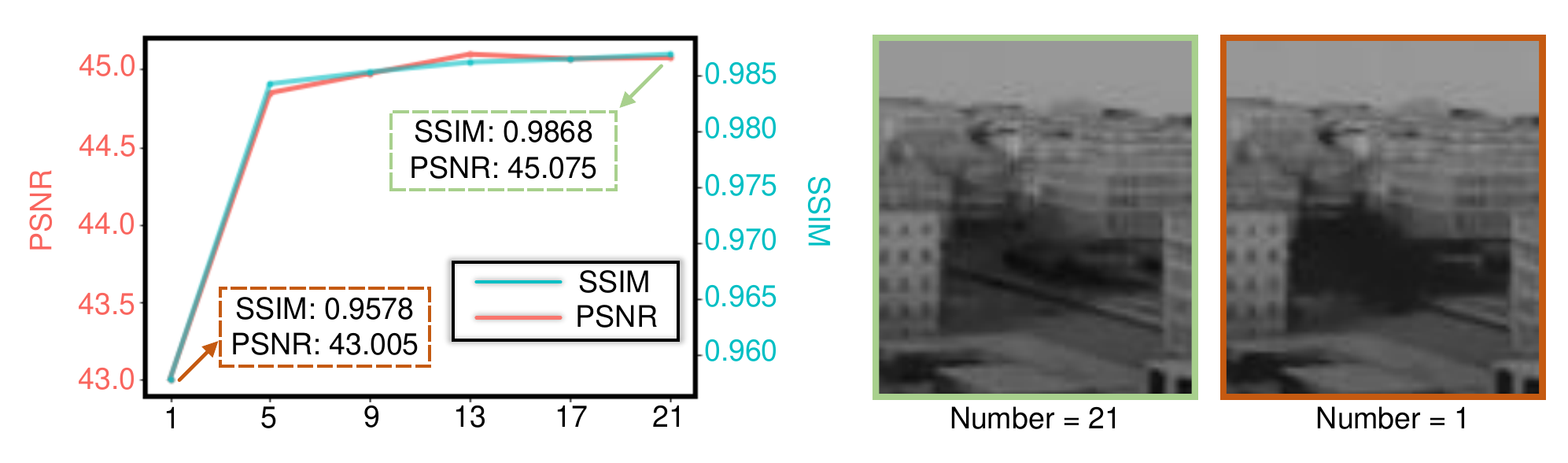}
\centering
\caption{Effect of the number of continuous spike streams on the performance. We test on the dataset, LLR. \textbf{Left}: PSNR and SSIM of the method under the different number of continuous spike streams. \textbf{Right}: Comparison of reconstruction images.}\label{number}

\end{figure}

\hupar{The number of continuous spike streams} For solving the reconstruction difficulty caused by inadequate information in low-light scenes, the release time of spike in LR-Rep and temporal features in fusion module are maintained forward and backward in a recurrent manner.  The number of continuous spike streams has an impact on our method performance. Fig.~\ref{number} shows its effect on the performance. We can find that, as the number increases, the performance of our method can greatly increase until convergence. This is because, as the number increases, our method can utilize more temporal information until sufficient. The reconstrued image from 21 continuous spike streams has more details in a shaded area.

\section{Conclusion}
We propose a bidirectional recurrent-based reconstruction framework for spike camera to better handle different light conditions. In our framework, a light-robust representation (LR-Rep) is designed to aggregate temporal information in spike streams. Moreover, a fusion module is used to extract temporal features. To evaluate the performance of different methods in low-light high-speed scenes, we synthesize a reconstruction dataset where light sources are carefully designed to be consistent with reality. The experiment on both synthetic data and real data shows the superiority of our method.



\hupar{Acknowledgement} This work was supported by the National Science and Technology Major Project (Grant No. 2022ZD0116305), the Beijing Natural Science Foundation (Grant No. JQ24023), and the Beijing Municipal Science \& Technology Commission Project  (No.Z231100006623010).


%
%

\end{document}